\documentclass[letterpaper]{article} 
\usepackage{aaai2026}  
\usepackage{times}  
\usepackage{helvet}  
\usepackage{courier}  
\usepackage[hyphens]{url}  
\usepackage{graphicx} 
\usepackage{natbib}  
\usepackage{caption} 
\usepackage{algorithm}
\usepackage{algorithmic}
\usepackage{xcolor}
\usepackage{comment}
\usepackage[export]{adjustbox}
\usepackage{graphicx}
\usepackage{subcaption}

\usepackage{fancyhdr}
\newcommand{\name}{{SpiderGen}\xspace}

\newcommand{\graphname}{{PFG}\xspace}

\newcommand{\numsamples}{{65}\xspace}

\newcommand{\accuracy}{{65\%}\xspace}

\usepackage{caption} 
\usepackage{xspace} 
\usepackage{amsmath}

\usepackage{newfloat}
\usepackage{listings}
\DeclareCaptionStyle{ruled}{labelfont=normalfont,labelsep=colon,strut=off} 
\floatstyle{ruled}
\newfloat{listing}{tb}{lst}{}
\floatname{listing}{Listing}
\title{\name: Towards Procedure Generation for Carbon Life Cycle Assessments With Generative AI}

\author {
    Anupama Sitaraman\textsuperscript{\rm 1},
    Bharathan Balaji\textsuperscript{\rm 2}\footnote{Work unrelated to Amazon.},
    Yuvraj Agarwal \textsuperscript{\rm 1}
}
\affiliations {
    \textsuperscript{\rm 1}Carnegie Mellon University\\
    \textsuperscript{\rm 2}Amazon\\
    asitaram@andrew.cmu.edu, bhabalaj@amazon.com, 
    yuvraj@cs.cmu.edu
}

\usepackage{bibentry}

\begin{document}

\maketitle

\begin{abstract}
Investigating the effects of climate change and global warming caused by GHG emissions have been a key concern worldwide. These emissions are largely contributed to by the production, use and disposal of consumer products. Thus, it is important to build tools to estimate the environmental impact of consumer goods, an essential part of which is conducting Life Cycle  Assessments (LCAs). LCAs specify and account for the appropriate processes involved with the production, use, and disposal of the products. We present \name, an LLM-based workflow which integrates the taxonomy and methodology of traditional LCA with the reasoning capabilities and world knowledge of LLMs to generate graphical representations of the key procedural information used for LCA, known as Product Category Rules Process Flow Graphs (PCR PFGs). We additionally evaluate the output of \name by comparing it with \numsamples real-world LCA documents. We find that \name provides accurate LCA process information that is either fully correct or has minor errors, achieving an F1-Score of \accuracy across 10 sample data points, as compared to 53\% using a one-shot prompting method. We observe that the remaining errors occur primarily due to differences in detail between LCA documents, as well as differences in the ``scope" of which auxiliary processes must also be included. We also demonstrate that \name performs better than several baselines techniques, such as chain-of-thought prompting and one-shot prompting. Finally, we highlight {\name}'s  potential to reduce the human effort and costs for estimating carbon impact, as it is able to produce LCA process information for less than \$1 USD in under 10 minutes as compared to the status quo LCA, which can cost over \$25000 USD and take up to 21-person days. 
\end{abstract}

\begin{links}
    \link{Code}{https://github.com/synergylabs/SpiderGen}
\end{links}

\label{sec:intro}
\section{Introduction}
\begin{figure}
    \centering
    \includegraphics[width=0.85\linewidth]{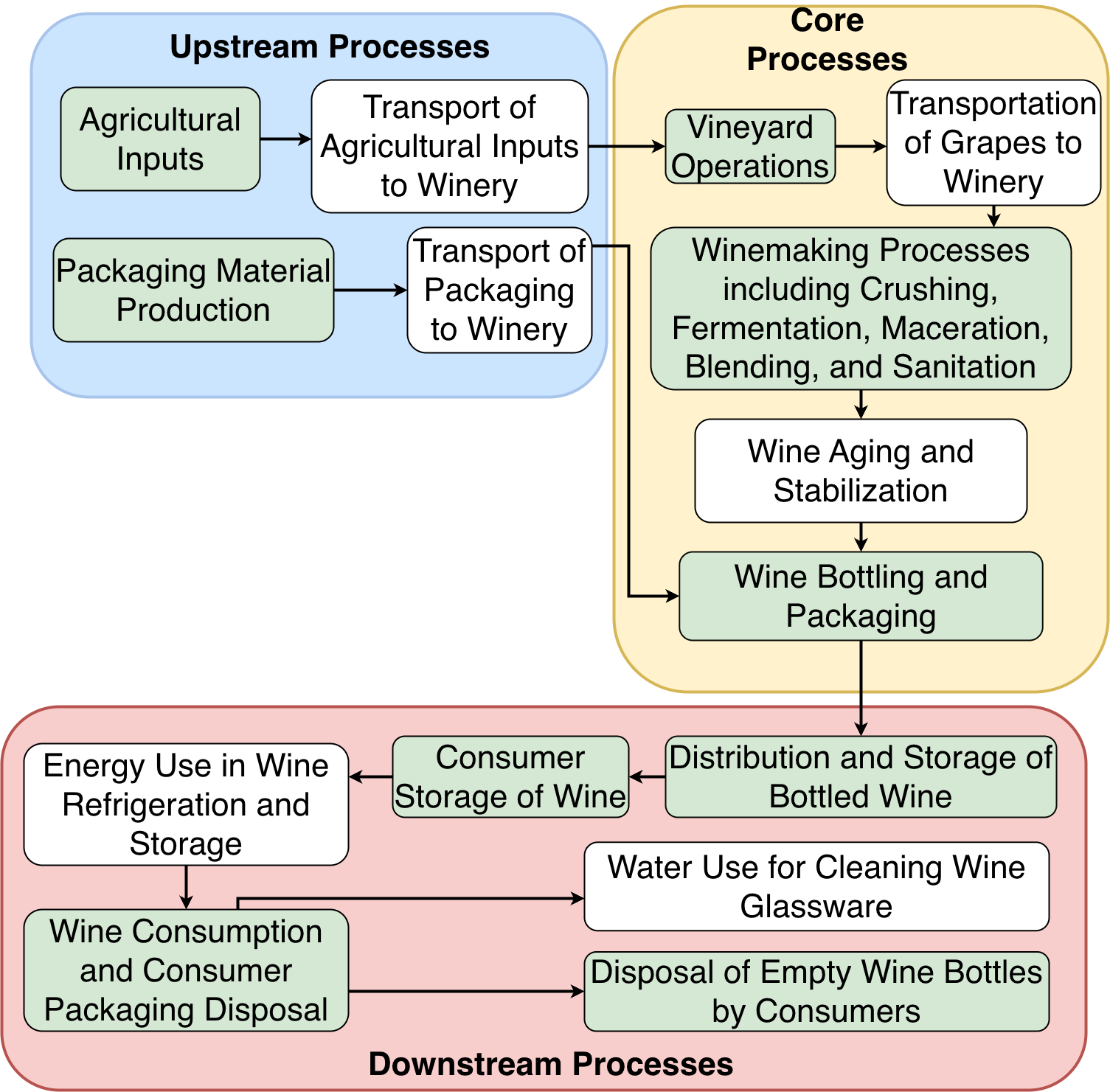}
    \caption{A simplified example \graphname for the product category ``Wine'' produced by \name.}
    \label{fig:wine}
\end{figure}
\begin{figure*}
\centering
    \includegraphics[width=\linewidth]{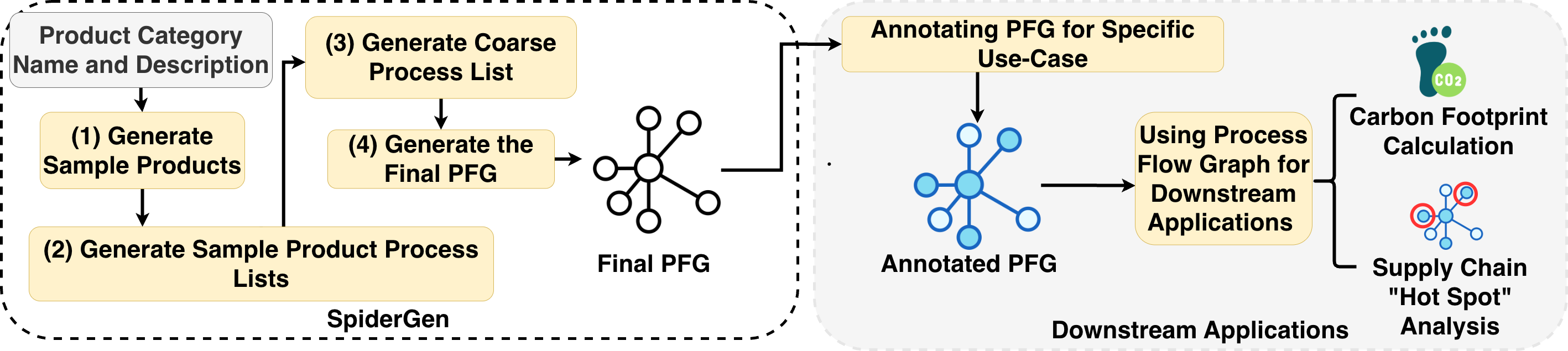}
    \caption{The \name workflow utilizes LLMs and sentence transformers to create Product Category Rule Process Flow Graphs (PCR PFGs). This workflow consists of 
    (1) A product generation step, where sample products are sourced and (2) processes for those sample products are generated, (3) A coarse process generation step, where the sample product processes are coarsened to create generalized processes and (4) the generation of the process flow graph itself. The formation of these graphs can then be used for a variety of applications, such as generating carbon footprints and and conducting supply chain analysis.} 
    \label{fig:system_diagram}
\end{figure*}

Life Cycle Assessments (LCAs) are commonly used to evaluate the environmental impact of a product in all phases of the product's ``life'', such as the manufacturing phase or the use phase by the end consumer.
Although LCAs are essential for measuring the environmental impacts of products, the end-to-end time and monetary cost of conducting LCAs is extraordinarily high - a 2017 study shows that just determining the appropriate processes for a given product category can cost over \$25000 USD, and can take up 21 person days to produce~\cite{tasaki_international_2017}. Due to the challenges of generating the appropriate processes for an LCA, it is very difficult to scale environmental impact assessments. As a result, LCAs are available for a very small subset of products.

A key challenge in creating an LCA is choosing which processes to include in the analysis, and creating standardized methodologies with which to evaluate specific categories of products. Traditionally, the appropriate processes for a given product category are determined by a committee of experts, and the LCA practitioner utilizes these processes and augments them for the specific product. These processes are described using a Process Flow Graph (\graphname) to represent the ordering and dependencies throughout the value chain of each product. An example of a \graphname for the product category ``Wine'' is shown in Figure \ref{fig:wine}. 
To address this challenge, there has been a growing interest in using machine learning techniques to automate LCA procedures to allow for easier, faster and more ubiquitous LCAs. Prior work has primarily focused on using existing process rules for specific categories of products, ``bill of materials'' (BOMs), and LCA guidelines, and automating other parts of the LCA pipeline \cite{balaji_caml_2023, balaji_flamingo_2023, balaji_parakeet_2024, zhang_deltalca_2024, sousa_approximate_2002, Wang2025}. However, to our knowledge, we are the first to explore the potential of using ML to generate {\graphname}s for LCA. Automating this key step will enable LCAs of products which have no guiding \graphname available previously, such as toothbrushes. This problem is complex, as it requires the collection of a wide range of process information, understanding the dependencies of these processes, and the ability to generalize processes to be applicable to all products of a given category.

\textbf{Our Approach.} We propose \name \footnote{Much like a spider, \name is a workflow which generates a web of resources and weaves insights together to generate {\graphname}s.}, a novel LLM-based workflow that automates the production of {\graphname}s for a given product category. To do so, \name
utilizes the world knowledge capabilities of LLMs, as well as text-embedding methods such as SBERT \cite{reimers-2019-sentence-bert}, and graph clustering algorithms such as K-Means clustering to derive upstream, core, and downstream processes for the overarching product category. \name then orders these processes to produce a Directed Acyclic Graph (DAG) representation of the end-to-end life cycle of the product category. Notably, we leverage specific \textit{ontologies} and \textit{taxonomies} involved in LCA production \cite{EPD, epa_pcr_us_2024}, to be able to generate relevant processes that are required in the \graphname. We specifically address the challenges of generating processes that are \textit{generalizable} across all products of a given product category, ordering them based on their dependencies, and ensuring that all of the processes that may be within the ``scope" of a true LCA are included. 

\textbf{Evaluation Methodology.} To evaluate \name, we utilize a ground truth dataset of \numsamples LCA documents from EPD International \cite{EPD}. 
To compare the generated {\graphname}s to the ground truth, we introduce a novel evaluation methodology (Section \ref{sec:eval_metric}) since prior knowledge graph evaluation methods, such as those for UML diagrams and Business Process Flow graphs are not appropriate for our problem \cite{fauzan_structural_2024, faisal_approach_2018, kristina_business_2024}.
Complete {\graphname}s must include all the correct processes and nothing more, and must include the correct partial ordering of processes, which reflect the dependencies of each process. To determine whether \name has generated the correct processes and ordering, we provide both a qualitative and quantitative evaluation of \name. For the quantitative evaluation, we calculate the Pointwise Mutual Information (PMI) \cite{xu_benchmarking_2025} between the ground-truth and {\name}'s generated {\graphname}s for all {\numsamples} LCA documents. For qualitative evaluation, we outline a set of criteria that determine ``correctness" and evaluate 10 of the \numsamples generated {\graphname}s using this criteria. 

\textbf{Our Contributions.} \name serves as a key tool to determine the environmental impacts of many products.  We make the following contributions:
\begin{itemize}
    \item We describe and implement \name, a novel zero-shot workflow for producing {\graphname}s which utilizes the world-knowledge and text processing capability of LLMs.  
    \item We introduce an evaluation methodology for comparing {\graphname}s using Pointwise-Mutual Information estimation, and conduct a qualitative analysis of {\name}'s ability to produce the necessary processes for LCA. 
    \item We implement and apply the \name pipeline to generate {\graphname}s using the United Nations Central Product Classification (UNCPC) descriptions and evaluate the generated {\graphname}s using real-world LCA procedure documents for general product categories.
    \item We show that \name accurately produces {\graphname}s relative to the ground truth, yielding an average F1-score of \accuracy, as compared to an average F1-score of 53\% using a one-shot prompting method. We find that the main challenges of increasing the accuracy of \name lie in producing all of the appropriate auxiliary processes that are part of the ground truth \graphname, while avoiding extrapolated processes that may not be applicable to a given product category. 
\end{itemize}

\label{sec:related_work}
\section{Related Work}
\subsection{Traditional LCA Frameworks}
LCA is the method used to estimate the carbon footprint of products and processes. A LCA expert starts by identifying the processes and environmental factors of the product life cycle that must be included. To do so, LCA experts rely on Process Flow Graphs 
({\graphname}s), which describe the processes for a specific product category that must be included in the LCA and their dependencies. By utilizing these graphs, experts can avoid making errors while conducting LCAs. Further, these graphs allow products of the same category to be evaluated in a consistent way, allowing for appropriate LCA comparisons. Additionally, these graphs also prevent ``greenwashing", where environmental impact information is manipulated to make products appear more sustainable than they truly are \cite{brandao_life_2024}.

{\graphname}s are officially made available within Product Category Rules documents, and are created by a team of industry experts. However, the process of creating {\graphname}s is time and resource intensive, making them inaccessible. For example, although {\graphname}s exist for many food products, they do not exist for computing technology products, such as laptops. Due to the lack of accessible LCA information, there is growing interest in automating LCA using ML.

\subsection{Machine Learning for LCA}
Incorporating machine learning techniques for different parts of the LCA process has become a growing area of interest \cite{algren2021}.
Some of the earlier efforts utilize surrogate modeling to create new LCA analyses based on a training set of LCAs for a set number of products \cite{sousa_approximate_2002, sousa_learning_nodate, sousa_product_2006}. 
More recently, researchers have proposed using machine learning and LLMs to extract relevant information from LCA reports and databases~\cite{goridkov_whats_2024}.  
 Balaji et al. use sentence embeddings and LLMs to match environmental impact factors from LCA databases to a given product descriptions \cite{balaji_flamingo_2023, balaji_caml_2023, balaji_parakeet_2024}. A related body of work uses information extraction methods to \textit{synthesize} LCAs by utilizing templates for a given product, such as a Bill of Materials (BOM) and domain-specific databases to guide the extraction, synthesis or prediction of LCAs \cite{zhang_deltalca_2024}. In contrast to our work, these prior works do not explore the generation of {\graphname}s for LCA.

\subsection{Procedural Generation Using LLMs}
Using LLMs to generate procedural information has also become popular in different domains such as generating food recipes, UML diagrams, and Business Process Flow diagrams. The approaches either generate new procedures using a fine-tuned language model \cite{Mohbat_2024}, or extract structured outputs from existing procedure documents. The most closely related work to ours utilizes LLMs to structure the outputs of existing Standard Operating Procedures in the case of recipe generation and Business Process Flow diagrams \cite{garg2025generatingstructuredplanrepresentation}. While this work organizes existing procedural information into a standard operating procedure graph, our approach contains an additional step of generating the processes themselves.

\subsection{Our Approach}
To the best of our knowledge, we are the first to explore the problem of generating process flow graphs for LCA. In contrast to prior work, our approach is zero-shot and uses off-the-shelf models to generate {\graphname}s. We achieve this by applying the ontology and taxonomies of LCA production to create an LLM workflow for graph construction for our specific problem. Unlike fine-tuning based methods, \name is not constrained by limited training data, which allowing \name to create a {\graphname} for any product category.

\label{sec:system_overview}
\section{Problem and Preliminaries}
We introduce \name, an LLM-based workflow which generates a graphical representation of the life cycle processes of a given product, called a Process Flow Graph (\graphname). 
We define the \graphname for a given product category $pc$ to be $G_{pc} = (V_{pc}, E_{pc})$, where the vertices $V_{pc}$ represent the \textit{generalized} processes for a given product category, and the edges $E_{pc}$ represent their ordering.
$G_{pc}$ generation uses the \textit{ontology} and \textit{taxonomy} of LCA \graphname construction. The nodes in $V_{pc}$ span three life cycle phases: ``upstream" processes include all life cycle processes that come before the manufacturing of the product (ex. raw material extraction), ``core" processes involve all manufacturing steps, and ``downstream" processes include processes that come after product manufacturing (ex. consumption). To reflect this taxonomy, each node in $V_{pc}$ is given a label for the life cycle phase during which it occurs (either ``upstream", ``core" or ``downstream").
In our setting, $G_{pc}$ is a directed acyclic graph (DAG) and is made up of two sets of edges: $E_m$ and $E_s$, such that $(E_{pc} = E_m \cup E_s$). $E_m$ indicates the ordering between main processes (i.e, for edge $(i,j) \in E_m$, process $i$ \textit{precedes} process $j$). $E_s$ indicates the relationship between sub-processes and main processes (i.e., for edge $(i,j) \in E_s$, $i$ is a \textit{subprocess} of j). Additionally, we require that all upstream processes must precede core processes, and all core processes must precede downstream processes.
\section{\name: Our Proposed Solution}
We now describe how \name produces $G_{pc}$. Generating $G_{pc}$ involves the following steps (as shown in Figure \ref{fig:system_diagram}).

\textbf{(1) Generate Sample Products:} Since $G_{pc}$ is generalizable across different products in a given category, we generate a list of real products that are relevant to this category. The goal of this step is to get a lay of the land of common processes required for products in this category, drawing this information as much as possible from real examples. There are two challenges with generating sample products: 

\textit{Ensuring Product Diversity}: The listed products must ideally use a diverse set of processes and raw materials to avoid the issue of overfitting to an overrepresented type of product within a product category.

\textit{Quantity}: To gain a sufficient understanding of a product category, we must ask for the right number of products to use. In early experiments, we found that asking for too few products would result in \graphname information that was over-fitted to a specific niche within a product category. However, querying for too many products may make the \graphname over-generalized and therefore inaccurate.

\textbf{(2) Generate Sample Product Process Lists:} To find common processes across a given product category, \name instructs an LLM to select diverse products that use a wide variety of materials and processes. For each product, \name queries an LLM for details about what components make up the product, and how these components are processed. We then query an LLM to find the processes involved in manufacturing and distributing the product. We prompt an LLM to list the process name, the life cycle phase it is involved in, and describe why this process is included. We use the LCA ISO standards to guide the LLM for process generation to cover various types of processes and ensure complete coverage of environmental impacts \cite{EPD}.

\textbf{(3) Generate Coarse Process List:} From the detailed process descriptions for a list of products, we create coarser generalized processes that can be included in a \graphname. Intuitively, similar processes that appear in all product processes should be included in the {\graphname}s, since it is likely that they are an important process to include for the entire category. Based on this intuition, we identify clusters of common processes to include in the final \graphname. Finally, we utilize an LLM to summarize these clusters to create coarse process descriptions and eliminate clusters that are repeated. 
 To generate relevant clusters of nodes, we utilize a pre-trained SBERT model to embed each process. We then use K-means clustering to provide groupings of repeated processes. To ensure that we do not cluster processes that may seem similar but are a part of different life cycle phases within the final {\graphname}, we only cluster processes that are in the same life cycle phase. For each life cycle phase, we select the number of clusters that minimizes the Davies Bouldin score of each group life cycle phase clusters. Using these clusters, we then prompt the LLM to give a description of each of the clusters and then remove repetitions. The resultant descriptions are the coarse processes used in the final process flow graph.
 
\textbf{(4) Generate the final \graphname:} Given the processes that form the vertices of $G_{pc}$, we now generate the edges of the graph to place these processes in the correct ordering. 
We establish two types of ordering: \textit{explicit} ordering, 
and \textit{implicit} ordering. 
Implicit ordering is based on the life cycle phases associated with each process. For example, processes that are a part of the ``upstream'' phase will always appear before processes that are a part of the ``core'' phase. After creating these implicit orderings, we use an LLM to generate explicit orderings within processes with the same life cycle phase. For example, the LLM will reason that a process called ``Mixing dough'' will appear before ``Forming noodles with dough'', where both are core processes. From our observation, modern LLMs can reason about explicit ordering quite well, and provide orderings that align with real process orderings. By using implicit ordering based on life cycle phases dictated by the taxonomy of LCA, \name is able to generate more accurate explicit orderings within a smaller set of processes that belong to the same life cycle phase.  
Using this method, we observe that \name is able to create process orders that are close to the correct {\graphname}s.


\label{sec:eval_metric}
\section{Evaluation Methodology}
\label{sec:eval_setup}
\subsection{Evaluation Setup}
We evaluate the generation of \graphname $G_{pc}$ on a set of \numsamples Product Category Rules (PCR) documents in the EPD International Database \cite{EPD}. These documents contain generalized processes for a given product category or a set of product categories for each phase of the product life cycle (upstream, core, and downstream). We manually extract the processes from these PCR documents and form a \graphname with these processes.
Each of the PCR documents have United Nations Central Product Classification (UN CPC) codes that describe the product category. For each PCR, we compile descriptions of the relevant UN CPC codes, as well as the name of the PCR. This UN CPC information is given to \name as an input for generating the \graphname. We experiment with three different LLM models: OpenAI's GPT-4o, o1-preview and o1-mini models. We used these models for our evaluation from April to July 2025. 

\subsection{Baseline Methods for Evaluation}
As there are no existing evaluation baselines for our problem, we present two baseline methods which draw upon existing techniques in LLM prompt engineering:

\textbf{LLMDirect:} The LLMDirect method directly prompts the LLM for the {\graphname} using the well-established method of \textit{Chain-of-Thought Reasoning} \cite{sahoo2025systematicsurveypromptengineering}. With this method, we provide the LLM with step-by-step instructions on how to gather all the necessary processes to create the {\graphname}, as well as how to create the graph using these steps. We also prompt it to provide a rationale for each step.We include our exact prompt for this method in the Technical Appendix.

\textbf{LLMExample:} The LLMExample method is an improvement on LLMDirect and involves the developer conducting \textit{One-Shot Prompting} \cite{sahoo2025systematicsurveypromptengineering}, where the LLM is given an example {\graphname}, and must follow this example to create a new {\graphname} for an entirely different product category.  
\subsection{Qualitative Evaluation Methodology}
We qualitatively analyze the results for 10 different product categories by comparing the {\graphname}s generated by \name to the ground truth {\graphname}s, categorizing any errors that \name makes. This procedure involves two steps:
We first pair all of the nodes by similarity by pairing every node from the generated graph $G_{pc}$ to the most similar node from the ground truth graph. Every node from both graphs must be included in the matching, and many-to-many matching are allowed. If there is no appropriate pairing for a given node, we pair the node with a fake node called ``N/A". 

We then evaluate the matches using the following criteria:
\begin{itemize}
\item If there exists a match such that the ground truth node corresponds to that in the generated graph, we label this pairing as a ``match". Note that we are considering cases where part of the process label is exactly matching to be labeled as ``match'' (i.e, if there is a ground-truth node called ``Agriculture, including electricity and water consumption” and there is a generated node called ``electricity consumption of agricultural processes”, this is a ``match”).
\item If there exists a generated node that is an inferred (but not explicitly stated) sub-process of a ground-truth node, this match is labeled as ``subprocess'' 
(i.e, if there is a ground-truth node called ``Manufacturing Pasta'' and a generated node called ``Cutting pasta dough'', this is a ``subprocess”).
\item If there exists a generated node that is a specific version of a process stated in the ground-truth graph, this match is labeled as ``specific''. (i.e, if there is a ground-truth node called ``Cultivating grains for pasta'', and there is a generated node called ``cultivating durum wheat for pasta'', this is a “specific process”). 
\item If there exists a generated node which does not have an appropriate pairing with a ground-truth graph node, it is labeled as ``wrong'', because in these scenarios, \name has hallucinated a step that does not exist for the product category. If there exists a ground-truth node which does not have an appropriate generated node pair, it is labeled as ``missing'', because in these scenarios, \name has missed including the process. 
\end{itemize}
\textbf{F1 Scoring:} We calculate an F1 score to capture the overall quality of the generated graph. The F1 Score is defined as 
\begin{equation}
\label{eq:f1_score}
    (2*\text{Precision}* \text{Recall})/(\text{Precision} + \text{Recall})
\end{equation} 
where Recall is calculated as 
\begin{equation}
\text{Recall}= (n_{specific} + n_{subprocess} +n_{match})/(n_{groundtruth})
\end{equation}
and Precision is calculated as
\begin{equation}
\text{Precision}= 1- (n_{wrong})/(n_{generated})
\end{equation}
where $n_{specific}$ is the number of ground-truth nodes that are in ``specific" pairings, $n_{subprocess}$ is the number of ground-truth nodes that are in ``subprocess" pairings, $n_{match}$ is the number of ground-truth nodes that are in ``match" pairings, $n_{groundtruth}$ is the total number of ground truth nodes, $n_{wrong}$ is the number of generated nodes that are labeled as ``wrong", and $n_{generated}$ is the total number of generated nodes.We provide a set of ten manually evaluated
\footnote{We initially utilized LLM-as-a-Judge techniques to rapidly conduct multiple of these evaluations. However, we observed that even with the addition of example cases, and chain-of-thought prompting, LLM-as-a-Judge tended to estimate a higher recall than a human judge. Additional details of this analysis can be found in the technical appendix.
analyses, and describe themes in the erroneous processes produced by \name in Table 1. We observe that the majority of errors produced by \name come from two sources: (1) misalignment of an appropriate ``scope" for the processes that must be included (i.e, how many tangential or auxiliary processes should be included), and (2) extrapolation and assumptions about how a product might be used and what materials may be involved to create the product. } 

\subsection{Quantitative Evaluation Methodology} To evaluate node similarity between the generated and the ground truth graphs, we calculate the Pointwise Mutual Information (PMI) to determine the similarity between lists of processes \cite{xu_benchmarking_2025}. Pointwise mutual information is a commonly used measure to compare the similarity of two texts, and measures the probability of two texts co-occuring by chance. In the case of text-comparison tasks, PMI is defined as

\begin{equation}
\label{PMI}
\begin{aligned}
PMI(X=x, Y=y) &= \log \Pr(Y=y \mid X=x)  \\
              &\;\; - \log \Pr(Y=y)
\end{aligned}
\end{equation}

where $X$ and $Y$ are two texts.

In our case, we calculate the PMI across two lists of processes, one from the \name generated $G_{pc}$, and the other from a ground truth LCA document . We consider $X$ to represent the random variable for the ordered list of processes generated by \name, and $Y$ to be the random variable representing the ordered list of ground truth processes.

Prior work indicates that using LLM-weights to calculate PMI can lead to improved evaluations, as this metric can be more sensitive to semantic degradations when comparing the semantic similarity of two texts \cite{xu_benchmarking_2025}. Thus, we closely follow this methodology from prior work. We utilize the Llama-8b-Instruct model weights for all texts for consistency, and calculate a normalized PMI using the LLM weights as our quantitative evaluation metric.

For our quantitative analysis, we follow the steps below:
\begin{itemize}
    \item \textit{Pre-processing both lists:} We organize both lists such that upstream processes are listed first, core processes second, and downstream processes third. Both lists start with ``raw material procurement'' processes, and end with ``end-of-life'' processes.  For ground-truth documents, direct references to other documents, links, references to other sections of the text and references to calculations have been removed. Additionally, ``optional" processes are marked as such. Xu et.al propose ``pre-processing" text by using an LLM to rephrase the texts before calculating PMI \cite{xu_benchmarking_2025}. However, this pre-processing is not feasible for our problem as it may increase the risk of introducing hallucinations that change the meaning and structure of the ground-truth text. 
    \item \textit {Generating Log Probabilities:} To generate the log probabilities for $\log{\Pr(Y = y| X = x)}$, we create a combined text block, first listing the ground-truth node list and subsequently adding the generated node list. We then tokenize the entire text block and generate the probabilities for the sequence of text using the weights of Llama-8b-Instruct. We then calculate the log of these probabilities. Similarly, we calculate $\log{\Pr(Y=y)}$ by tokenizing the list of ground-truth nodes and getting the probabilities via the weights of Llama-8b-Instruct. By calculating both $\log{\Pr(Y = y| X = x)}$ and $\log{\Pr(Y=y)}$), we can calculate the PMI of any two lists.
    \item \textit{Normalizing the PMI:} To normalize the PMI, we calculate the maximum possible PMI of the list of nodes. The maximum possible PMI occurs when the exact same text is generated twice. Thus, we calculate $PMI(x, y)/PMI(y,y)$, given that $y$ is the ground-truth node list, and $x$ is the generated node list. 
\end{itemize}

Using this method, we are able to provide a quantitative evaluation whether the content of the process lists are similar between ground-truth process lists and {\name}'s generated process lists. 

\label{sec:eval}

\begin{table}[t]
\label{table:qualitiatve_quantitative_results}

\begin{tabular}{c|c|c|c}\hline
ID &Product Category & PMI & F1-Score  \\
\hline
$c_1$ & Railways (44)  & 0.02 &0.4                      \\
$c_2$  & Shower Enclosures (40)  &  0.02 & 0.73                         \\

$c_3$  & T-Shirts, Tops (28) & 0.03 & 0.77                  \\
$c_4$  & Moka Coffee (22)   &   0.02 & 0.66                    \\
$c_5$ &Dairy Products (22)  & 0.05&0.63                      
\\

$c_6$ & Graphite Products (20) & 0.04 & 0.61                      \\
$c_7$ &Detergents \& Washing (19) &  0.09  &0.59 \\ 
$c_8$ &Woven Fabric (17)                      &0.05& 0.84                      \\
 $c_9$ & Air Ducts (17)    &  0.07 & 0.65    \\
$c_{10}$ & Bottled Water (12) & 0.03& 0.59
\\

\hline
\end{tabular}

\caption{Results of our qualitative analysis for 10 products (Circled in Figure \ref{fig:node_size}). We order the products based on the number of nodes (in parenthesis). 
Note that PMI is the quantitative metric defined in Equation \ref{PMI}.}\label{fig:ten_sample_table}

\end{table}

\begin{figure}[t]

\centering
\includegraphics[width=0.45\textwidth]
{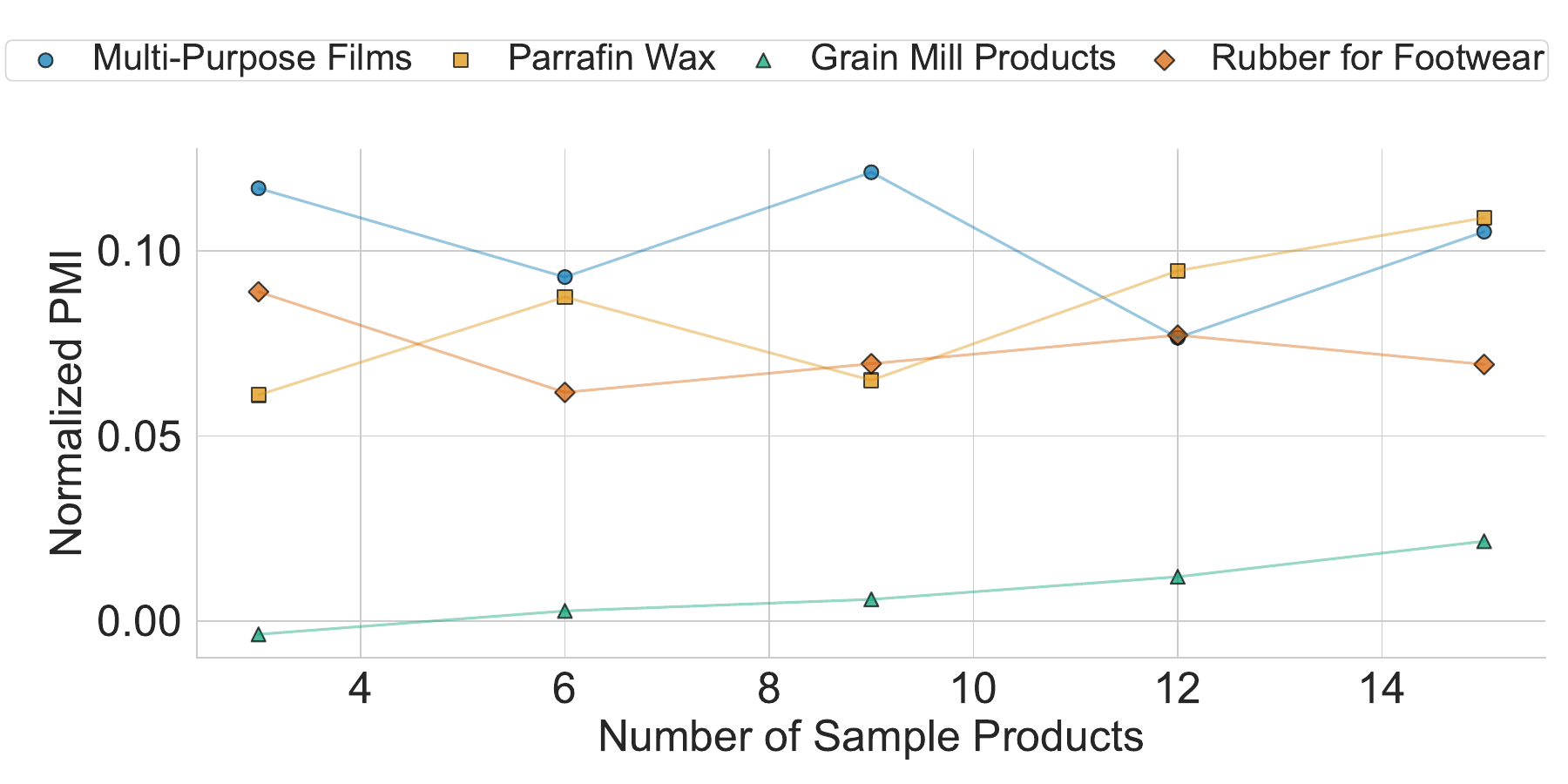}
        \caption{Comparing the normalized PMI values as different numbers of products in each category are generated in the first step of the \name workflow. We observe lower variability in PMI as the number increases.}\label{fig:num_samples}
\end{figure}
\begin{figure}[t]

\centering
\includegraphics[width=0.45\textwidth]{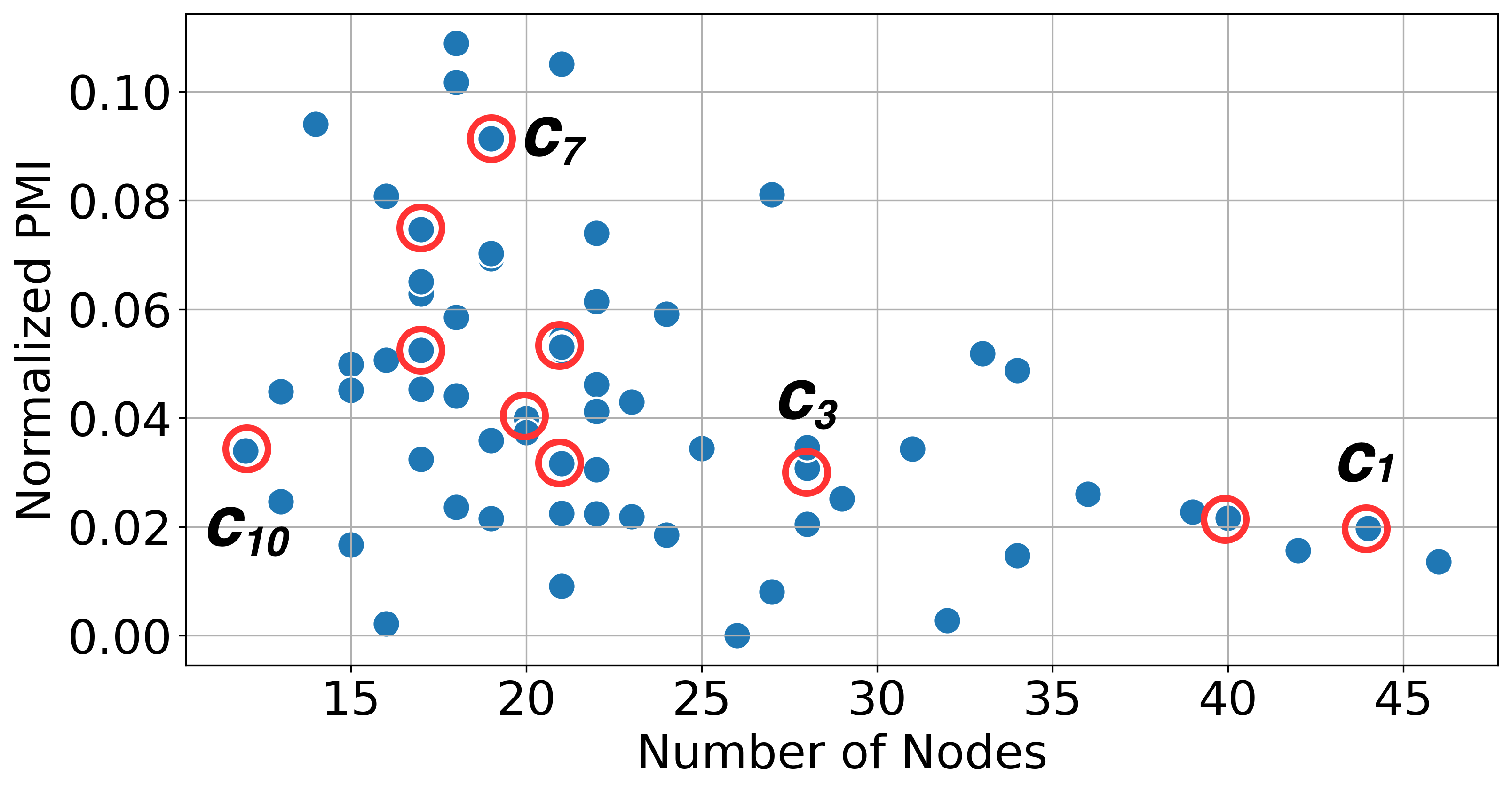}
        \caption{Comparing the Normalized PMI Scores for \numsamples  products with varying complexity as denoted by the number of nodes in the ground truth $G_{pc}$ graphs.  \name has higher normalized PMI for simpler product categories (e.g. ``Dairy Products") and lower for more complicated ones (e.g. ``Railways''). We also report the qualitative scores for 10 product categories (circled nodes) in Table \ref{fig:ten_sample_table}, and label some of the product categories in Table \ref{fig:ten_sample_table} in this figure.}\label{fig:node_size} 
       
\end{figure}
\section{Evaluation Results}

We present an evaluation of \name, highlighting key results. Additional evaluation results can be found in the extended version of our paper \cite{sitaraman2025spidergenproceduregenerationcarbon}.

\begin{table}[t]

\centering
\begin{tabular}{c|c|c}
\hline
Model  & Average  PMI & Average Cost per \graphname\\ \hline
o1-preview & $0.051  \pm 0.029$  & US \$$4.7 \pm 0.28$                      \\
o1-mini    & $0.046   \pm 0.023$   &     US \$$0.36\pm 0.03$                \\
gpt-4o     &$0.039    \pm          0.022  $  & US \$$0.64 \pm 0.14$ \\          \hline         
\end{tabular}

\caption{Average PMI Score and cost for \name using different LLMs across 40 product categories. \name performs better with models, such as OpenAI's o1-preview and o1-mini. o1-mini provides a good tradeoff with good performance and lowest cost.}

\end{table}
\subsection{Evaluating the Parameters of \name}
We first examine two key parameters that affect the \name workflow: (a) the selection of LLM model and (b) the number of example products used in the initial product generation step. 
\subsubsection{Efficacy vs Cost trade-offs of utilizing reasoning models:}
We implemented \name using 3 OpenAI models: gpt-4o, o1-mini, and o1-preview. o1-mini and o1-preview are reasoning models, with o1-mini being a smaller, faster, and cheaper model than the other two. Table 2 compares the performance of these models in terms of the average normalized PMI and cost for generating a single PFG across 40 product categories. The two reasoning models (o1-mini and o1-preview) outperform gpt-4o since they are able to provide more detailed outputs. However, o1-mini is close in terms of performance to o1-preview while being 1/13th the cost at less than US \$1 on average per \graphname. This cost is far lower than the typical production of a PFG, which can be upwards of US \$25000 \cite{tasaki_international_2017}.

  \subsubsection{Effect of the number of products generated by \name:}
 We evaluate the effect of producing a different number of sample products in the first step of \name (shown in Figure \ref{fig:num_samples} for four product categories). We evaluate the production of {\graphname}s using 3, 6, 9, 12, and 15 sample products for each \graphname generation. As shown in the figure, the PMI often increases with the addition of more sample products. We note that less ``niche" product categories, such as ``Grain Mills'' benefit from a larger number of sample products, and more ``niche" categories, such as ``Moka Coffee" benefit from smaller numbers of sample products, and can even have deteriorated {\graphname}s, as the number of sample products increases. Since generating 15 products works well for many scenarios, we use it for our implementation of \name. 
 
\subsection{Evaluating \name on Ground Truth Data}

To evaluate \name end-to-end, we execute our pipeline for \numsamples product categories and compare the generated {\graphname}s to the ground truth graphs extracted from real LCA documents\footnote{Due to limited available PCR documents, we solely utilize these \numsamples PCRs from EPD International \cite{EPD}} \cite{EPD}.

\subsubsection{Efficacy of \name compared to baselines:} We compared \name with the two baseline methods, LLMDirect and LLMExample, across \numsamples product categories. Our results show that \name exceeds both baseline methods, providing a median normalized PMI score of $0.043 \pm 0.026$ across different product categories as compared to $0.026 \pm 0.020$ for LLMDirect and $0.029 \pm 0.023$ PMI for LLMExample. Based on our qualitative analysis of a subset of ten products, we find that \name avoids generating overly specific processes and captures more non-obvious auxiliary processes as compared to the baselines. 

\textit{Limitations of LLMDirect:} Although a Chain-of-Thought process may work for many problems, we found that an algorithm such as LLMDirect is insufficient to solve our problem. Firstly, it is unable to capture all the nuanced steps that are necessary for complete {\graphname}s. LLMDirect results in frequently missed steps that are peripheral to the main manufacturing processes, such as downstream product maintenance and disposal steps. Further, \graphname is sometimes overly specific to one subcategory of the product category. 

\textit{Limitations of LLMExample:} While LLMExample is more effective than LLMDirect, since it is able to capture the nuanced steps of the \graphname by following an example, it is still insufficient to solve our problem. LLMExample misses peripheral steps, like ``machine maintenance," that may be important to a broader product category. Additionally, the \graphname is still overly specific to a subset of items of the product category. 

\subsubsection{\name on Varying Product Complexity:} {\graphname}s can be described as more complex if they have more life cycle processes. For example, ``railways" is the second-most complicated product category in our dataset, as the PCR for ``railways" includes 44 different processes that must be taken into account. To evaluate whether \name is able to produce these complex {\graphname}s, we evaluated \name on ground-truth {\graphname}s with varying levels of complexity. Figure \ref{fig:node_size} shows a scatter plot of the normalized PMIs of {\graphname}s with different numbers of nodes. We find an overall trend is that for products that require more nodes, the PMI is lower than with product categories with smaller {\graphname}s. Both of these factors indicate that \name is sensitive to a change in product complexity. As it stands, \name produces higher-quality {\graphname}s for simpler products.


\section{Discussion and Conclusion}
\label{sec:discssuion_conclusion}
In this work, we introduce \name, a novel machine learning workflow that automates the generation of {\graphname}s, and introduce evaluation methods for measuring the correctness of {\graphname}s. We present an implementation of \name, and provide an evaluation of \name across a set of \numsamples product categories. We find that \name is able to capture a large portion of required processes for tracking the carbon footprint of various product categories. 
\subsection{Limitations and Open Problems}

Although \name makes strides towards automating the generation of {\graphname}s, we highlight several limitations of our work that should be addressed to enable workflows like \name to be adopted in the real world.

\textit{Increasing transparency in LCA automation.} Recent work indicates that providing \textit{transparency} in automated LCA methods is a key criteria \cite{ulissi_criteria_2025}. Thus, in \name, we enable transparency by  providing sample products, evidence and rationales for processes, and deterministic clustering techniques for forming coarser processes. We hope that this enables human LCA experts to gain a better understanding of how \name generates {\graphname}s. However, more work must be done to achieve other evaluation goals, such as measuring and indicating the uncertainty of the \graphname generation, as well as enabling LCA experts to collaboratively improve the \graphname estimates. We hope to explore these possibilities in future work.

\textit{Enhancing System Boundaries of \graphname Generation.}
We note that a primary limitation of \name is derived from the scope, or ``system boundary", of the \graphname being in conflict with the system boundary defined appropriately for the product category. In the real world, the system boundary is human-defined. However, for our experiments, we made assumptions about the system boundaries based on ISO standards and encoded them into SpiderGen. We note that this is the root cause for \name missing processes or hallucinating processes. Although the precision of \name is relatively high (on average 81\%), \name occasionally produces erroneous processes due to sticking closely to system boundaries that may be inappropriate for a given product category. For example, a more common error was to include processes such as ``replacement", or ``quality control", even if they were not relevant for a given product category. Additionally, \name misses some the auxiliary processes which are not directly related to the product manufacturing, material composition or usage.We believe that exploring human-AI collaboration methods to make appropriate choices for system boundaries is a promising next step for future work. 

\textit{Evaluating {\graphname}s in the wild.} A primary challenge with evaluating a workflow such as \name in contexts where there is either no ground-truth available, or where there may be disagreements between experts. For example, even with the ground-truth PCRs that we utilized for our evaluation, there may be experts who disagree on which processes should be included in a \graphname for a given product category. An ongoing challenge for future work would be to address enabling expert-evaluation for LCA automation tools, such that expert-consensus is considered. 

\textit{Citing the Sources: Utilizing Retrieval Augmented Generation.} In future iterations of \name, we believe that Retrieval Augmented Generation (RAG), where LLM models can search resources such as the internet, will be a promising method for increasing the transparency and traceability of \graphname generation.

\textit{Downstream Applications:} In future work, we hope to further explore the potential of \name for automating LCA in a variety of contexts, such as for Environmental Product Declarations (EPDs), Material Flow Analysis, which studies the flow of materials through a system, such as an economy, or a manufacturing process.

\section*{Acknowledgments}
This work is supported by the National Science Foundation Award CNS-2325956, as well as the National Science Foundation Graduate Research Fellowship Program under Grant DGE-2140739. Any opinions, findings, and conclusions or recommendations expressed in this material are those of the author(s) and do not necessarily reflect the views of the National Science Foundation. 

\bibliography{sample-base}

\newpage
\appendix
\newpage
\section{Technical Appendix}

This section contains additional information about the \name workflow implementation, including prompts used for the workflow, as well as additional results.

\subsection{Additional Details About the Product Category Rules Dataset} We provide additional details about the dataset of product category rules that we utilize for our evaluation in Table 4. We note that  the product categories that we use in our dataset have a wide range of complexity, ranging from including just 12 processes to 46. We also note that we analyze products with varying levels of ``specificity", as indicated by the length of the shortest UN CPC code which corresponds to the product category, ranging from 2 digits (broadest product category) to 5 digits (most specific or niche product category).

\begin{figure*}
    \includegraphics[width=\linewidth]{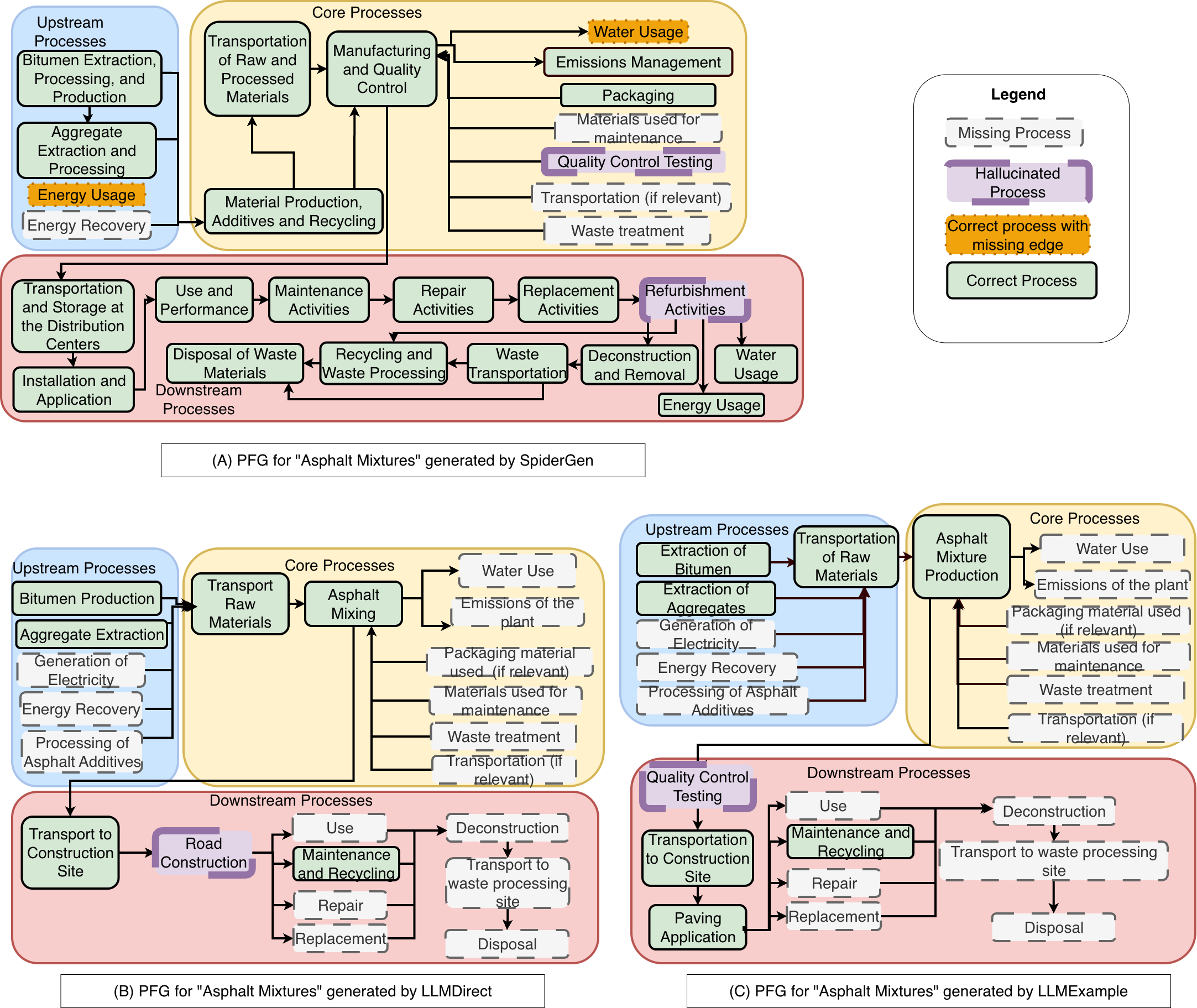}
    \caption{An example comparison of \graphname{}s between (a) \name, and two baselines, (b) LLMDirect and (c) LLMExample for the product category ``Asphalt Mixtures". We note that \name captures all downstream processes, and captures the majority of core processes, as compared to both LLMDirect and LLMExample, which fail to capture large portions of important processes, such as processing Asphalt additives, and maintaining asphalt.} \label{fig:asphalt_ex}
\end{figure*}

\begin{table*}

\begin{tabular}{c|c|c|c|c|c|c|c|c}\hline
Product Category & PMI & ROUGE-L & BLEU & BERTScore & Precision & Recall &F1-Score & LLM-as-a-Judge \\
\hline
Railways (44)  & 0.02 & 0.08 &0.0&0.74&1&0.25 & 0.4&0.68                  \\
Shower Enclosures (40)  &  0.02 &0.14&0.01&0.74 &0.9&0.63 & 0.74&0.73                       \\

T-Shirts, Tops (28) & 0.03 & 0.18 &0.05&0.8&1&0.63& 0.77&0.80                   \\
Moka Coffee (22)   &   0.02 &0.18&0.07&0.72 &0.7&0.64& 0.66&0.65                    \\
Dairy Products (22)  & 0.05 & 0.17 & 0.06& 0.84 &0.59&0.68&0.63&0.65                      
\\

Graphite Products (20) & 0.04 & 0.16&0.1&0.82&0.63&0.61&0.62&0.84                        \\
Detergents \& Washing (19) &  0.09 & 0.15 &0.05&0.73&0.6&0.57&0.59&0.55 \\ 

Woven Fabric (17) &0.05&0.17&0.1&0.81&1&0.73& 0.84&0.71                     \\
  Air Ducts (17)    &  0.07 & 0.13&0.28&0.72&0.71&0.59& 0.65&0.65   \\
Bottled Water (12) & 0.03&0.23&0.2&0.76&1&0.42& 0.59&0.83
\\

\hline
\end{tabular}

\caption{Results of \name for for 10 products (Shown circled in Figure 3) across multiple different metrics. We order the products based on complexity denoted by the number of nodes (in parenthesis). The F1-score, precision and recall are all provided by the qualitative analysis measure provided in Section \ref{sec:eval_metric}. We note that PMI is the most closely correlated to recall out of all evaluation metrics, which is why we utilize this metric throughout the paper. We additionally note that the LLM-as-a-Judge metric typically provides a higher score than the human-generated F1-score due to observing more of the generated processes matched to ground truth processes. }\label{tab:10_prod_analysis_more_detail}  

\vspace{-0.2cm}

\end{table*}

\subsection{Additional Evaluation Results for \name on Ground Truth Data}
In this section, we provide additional evaluations of \name on our ground-truth dataset of PFGs. 
\subsubsection{Comparing the \graphname{}s between both baselines and \name}Figure \ref{fig:asphalt_ex} shows a comparison between \name, and two baselines, LLMDirect and LLMExample, which are described in Section \ref{sec:eval_setup}. We note that in this case, LLMDirect and LLMExample miss significantly more processes compared to \name, which misses just four. We additionally note that \name primarily misses auxiliary processes that are not a part of the main procedures needed to create, use and dispose of asphalt. 
\subsubsection{Efficacy of \name in different life cycle phases:}

\begin{figure}
\label{fig:lifecycle}

\centering
\includegraphics[width=0.5\textwidth]{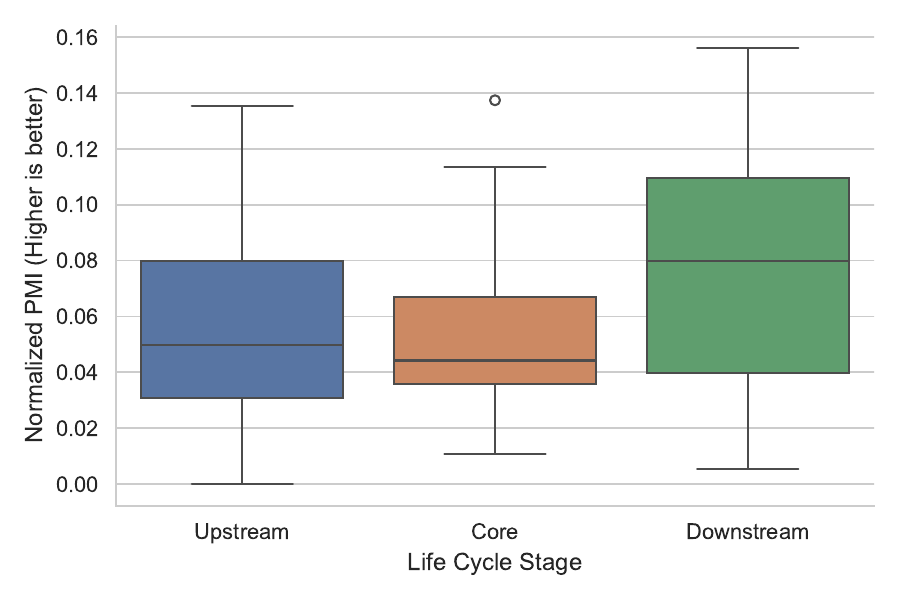}
        \caption{Normalized Mutual Information of \name results across different Life Cycle Phases. We observe the variance in performance is higher for "Downstream" processes than "Upstream" or "Core". }\label{fig:lca_stage_comparison}
\end{figure}

We explore the efficacy of \name in generating processes from different life cycle stages (either upstream, core, or downstream). Figure \ref{fig:lca_stage_comparison} shows the PMI scores of \numsamples different \name-generated {\graphname}s. We observe that \name demonstrates much lower interquartile ranges, as well as a lower median PMI score in upstream and core process determination than in downstream process determinations. 

We believe that the higher median PMI score for downstream processes is due to the fact that, unlike upstream and core processes which may be more niche to a specific product category, many of these downstream processes are standard across many product categories. For example, the process "disposal of product" or "waste management of product" will appear in a majority of product categories. This means that it is much easier for \name to predict downstream processes than other types of processes. We observe that \name typically generates as many of these standard downstream processes as they are relevant to the product category. For example,\name typically assumes all possible cases for the disposal of a product (ex. recycling, disposal, consumption, etc.).

However, it is often the case that downstream processes are \textit{under-specified} in real ground truth LCA documents. We observe that for a large portion of ground truth documents, the method of collecting downstream processes is either unclear or optional. This indicates that for specific product categories, there may be a higher level of uncertainty in downstream processes, and an LCA expert would be required to use their own interpretation of the document to conduct the LCA in this scenario. This increases the standard deviation of the PMI for downstream processes, as \name scores highly when compared to well-specified downstream process lists, and poorly when compared to optional or under specified downstream process lists. 

\subsubsection{Efficacy of \name on varying product specificity:} LCAs are conducted for product categories at varying levels of specificity. For example, there is a PCR for ``Grain Mill Products'', which is a broad category, and ``Disposable Surgical Drapes, Gowns, Air Suits and Face Masks'', which is a more niche category. Our aim in designing \name was to evaluate how well the system performs on product categories of differing specificity. In this evaluation, we define ``specificity'' using UN CPC codes, which are hierarchical codes. UN CPC codes have a larger number of digits (a maximum of 5) when the product category descriptor is more specific or niche, and a lower number of digits (a minimum of 2), when a product category descriptor is more broad. In our case, we measure the specificity of a ground-truth PCR to be the minimum number of digits of it's UN CPC code list.  Figure \ref{fig:specific} shows a box plot graph which shows the normalized PMI for more and less specific products. Based on these values, we can observe that \name accommodates broad categories best, with higher median PMIs for less specific product categories. We note that in our dataset, there are significantly fewer product categories that are at either extreme (with a CPC code length of 2 or with a CPC code length of 5), with just 3 product categories having a CPC code length of 2 and just 10 having a CPC code length of 5. 

\subsection{Additional evaluation methods}
We find that a core challenge of this work is to address how to adequately evaluate \name using evaluation metrics. In this section, we provide an evaluation of \name under additional evaluation metrics, and provide additional rationale for the utilization of PMI in the main body of the work over alternative metrics.

\subsubsection{Comparing PMI with traditional metrics} We first compare our PMI metric with metrics traditionally used to capture similarity, such as ROUGE-L, BLEU, and BERTScore. These comparisons are shown in Table \ref{tab:10_prod_analysis_more_detail}, We compare these scores to the qualitative evaluation scores (F1-Score, Precision and Recall), and find that PMI has the strongest correlation with recall out of all of the options for our dataset of 10 products.

\subsubsection{Utilization of LLM-as-a-Judge} In this section, we utilize an LLM-as-a-Judge technique to provide an evaluation of the correctness of the processes generated by \name. We implement LLM-as-a-Judge by listing out all \name-generated processes and all ground-truth processes in an LLM prompt. We first ask the LLM to determine whether each process that is generated by \name is included in the ground truth. From this, we are able to calculate \textit{precision}, where 
\begin{equation}
    precision = \text{num\_included}/\text{total\_num\_generated\_processes}
\end{equation}
We then ask the LLM the reverse question of which processes in the ground truth are given by the \name processes. From this, we are able to calculate \textit{recall}, where 
\begin{equation}
    recall = \text{num\_included}/\text{total\_num\_ground\_truth\_processes}
\end{equation}
We term the number of calculate the LLM-as-a-Judge score as an F1-score between this recall and precision value, which is calculated the same as in Equation \ref{eq:f1_score} in the main text.

The implementation details of our LLM-as-a-Judge are given in our GitHub repository (https://github.com/synergylabs/SpiderGen). We evaluate LLM-as-a-Judge across 10 products. We find that LLM-as-a-Judge routinely overestimates the recall in comparison to the human evaluator (in our dataset of 10 products, the LLM-as-a-Judge overestimated the F1-Score by 22.4\%). We find that the LLM-as-a-Judge method also underestimates precision by 11.9\% in this dataset of 10 products.

\begin{figure}[h!]
\label{fig:specific}
\includegraphics[width=0.5\textwidth]{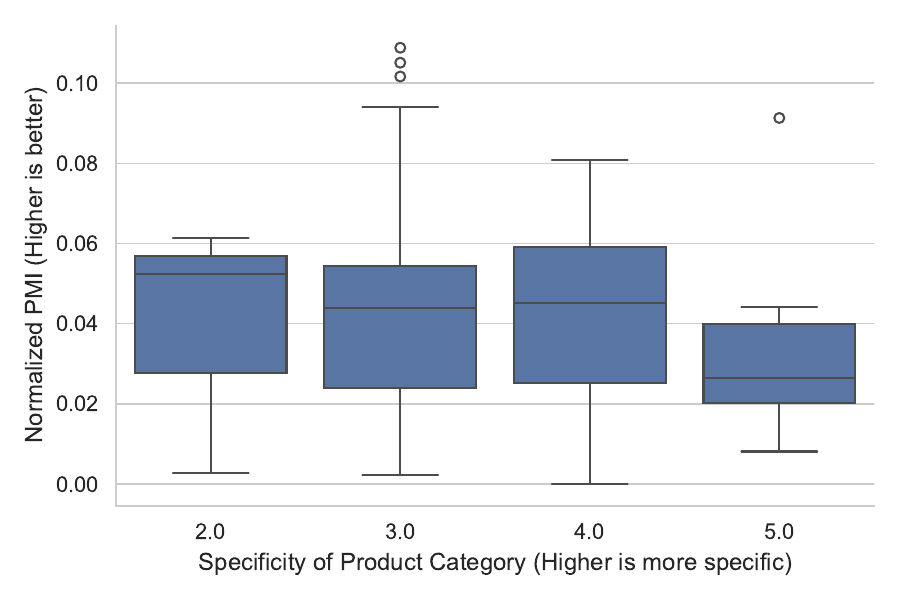}
        \caption{Normalized Mutual Information of \name results across different Product Specificity levels. We observe the standard deviation in performance is higher for broader product categories (such as those with a UN CPC code length of 4 or less), and is smaller for product categories that are the most specific (UN CPC code of 5). }\label{fig:specific}
\end{figure}

\vspace{-0.2cm}

\subsection{Implementation Details}
We implement \name in Python using, NumPy, Pandas, and the all-mpnet-base-v2 model from the HuggingFace Sentence Transformer Library \footnote{https://huggingface.co/sentence-transformers/all-mpnet-base-v2\#all-mpnet-base-v2}, as well as the KMeans function from the scikit-learn library \footnote{Pedregosa, F.; Varoquaux, G.; Gramfort, A.; Michel, V.;
Thirion, B.; Grisel, O.; Blondel, M.; Prettenhofer, P.; Weiss,
R.; Dubourg, V.; Vanderplas, J.; Passos, A.; Cournapeau, D.;
Brucher, M.; Perrot, M.; and Duchesnay, E. 2011. Scikit-
learn: Machine Learning in Python. Journal of Machine
Learning Research, 12: 2825–2830.}. We utilize OpenAI's API as well for making our api calls. 

For evaluating \name, we utilize the Llama-8b-Instruct model \footnote{https://github.com/meta-llama/llama3/blob/main/MODEL\_CARD.md} through HuggingFace Hub. We utilize the weights of this model. For further details on the use of the LLama-8b-Instruct model to calculate the PMI of two texts, refer to \cite{xu_benchmarking_2025}.

All code except for OpenAI model calls were run locally on an Apple MacBook Pro, with an M3 Pro chip (11-core CPU, 14-core GPU), with 18 GB RAM.  Further implementation details are provided in our GitHub repository (https://github.com/synergylabs/SpiderGen). 
\newpage
\begin{table*}[]
\small{
\hspace*{-0.2cm}
\begin{flushleft}
\begin{tabular}{lll}
Product Category & Number of Processes & UN-CPC Code Length\\ \hline
GRAPHITE PRODUCTS & 20 & 4 \\
OTHER SPECIAL- AND GENERAL-PURPOSE MACHINERY...& 21 & 3 \\
DISPENSING SYSTEMS & 20 & 4 \\
DAIRY PRODUCTS & 22 & 3 \\
BEARINGS, BEARING UNITS AND PARTS THEREOF & 21 & 4 \\
ELECTRICAL MOTORS AND GENERATORS AND PARTS THEREOF... & 18 & 5 \\
SKI FOOTWEAR & 15 & 4 \\
MEAT OF MAMMALS (FRESH, CHILLED OR FROZEN) & 24 & 4 \\
MACHINE-TOOLS FOR DRILLING, BORING OR MILLING... & 21 & 3 \\
MULTI-PURPOSE FILMS & 21 & 3 \\
SEATS & 16 & 4 \\
WOVEN, KNITTED OR CROCHETED FABRICS (OF SYNTHETIC FIBRES) & 
17 & 
3 \\
PARRAFIN WAX & 18 & 3 \\
GRAIN MILL PRODUCTS & 19 & 3 \\
RUBBER ARTICLES FOR FOOTWEAR & 19 & 4 \\
MOKA COFFEE & 22 & 5 \\
FISH, OTHERWISE PREPARED OR PRESERVED... & 21 & 4 \\
HIGHWAYS, STREETS AND ROADS & 39 & 4 \\
INDUSTRIAL FURNACES AND OVENS & 25 & 4 \\
WOVEN FABRICS OF SILK AND SILK-LIKE FIBRES & 33 & 4 \\
AIR DUCTS, SUBSTANTIAL MATERIALS (NON-CONSTRUCTION PRODUCT) & 17 & 4 \\
VEGETABLE JUICE, PLANT MILK, PLANT MILK BASED PRODUCTS... & 21 & 4 \\
SAUCES, MIXED CONDIMENTS AND MUSTARD & 19 & 5 \\
CLOSABLE FLEXIBLE PLASTIC PACKAGING & 18 & 4 \\
HOME AND SOHO GATEWAYS & 22 & 5 \\
PLASTICS IN PRIMARY FORM & 34 & 3 \\
CORROSION PROTECTION OF FABRICATED STEEL PRODUCTS & 27 & 5 \\
SHOWER ENCLOSURES & 40 & 5 \\
ASPHALT MIXTURES & 36 & 4 \\
TROUSERS, SHORTS AND SLACKS AND SIMILAR GARMENTS & 28 & 3 \\
TEXTILE YARN AND THREAD ...& 42 & 3 \\
FOOTWEAR & 23 & 3 \\
MEAT OF POULTRY (FRESH, FROZEN OR CHILLED) & 21 & 4 \\
RAILWAYS & 44 & 5 \\
PREFABRICATED BUILDINGS & 46 & 3 \\
T-SHIRTS, TOPS, SINGLETS AND OTHER VESTS & 28 & 3 \\
BASIC CHEMICALS & 18 & 3 \\
DETERGENTS AND WASHING PREPARATIONS & 19 & 5 \\
TEA & 22 & 5 \\
PARTS AND ACCESSORIES FOR COMPUTING MACHINES... & 17 & 4 \\
PASTA COOKED, STUFFED OR OTHERWISE PREPARED; COUSCOUS & 23 & 4 \\
PASSENGER CARS & 15 & 5 \\
FABRICATED STEEL PRODUCTS, EXCEPT CONSTRUCTION PRODUCTS... & 13 & 3 \\
DISPOSABLE SURGICAL DRAPES, GOWNS AND AIR SUITS & 19 & 4 \\
PREPARATIONS USED IN ANIMAL FEEDING FOR FOOD-PRODUCING ANIMALS & 16 & 3 \\
PORT OPERATION SERVICES & 17 & 3 \\
EVENTS AND TOURISM SERVICES & 32 & 2 \\
NONWOVEN WIPES & 13 & 5 \\
PULPS, WOODS OR OTHER FIBROUS CELLULOSIC MATERIALS & 24 & 4 \\
ARABLE AND VEGETABLE CROPS & 17 & 2 \\
TEXTILE MANUFACTURING SERVICES, NON-APPAREL FABRICS... & 28 & 3 \\
PRESERVES AND PREPARATIONS OF MEAT... & 16 & 4 \\
BAKERY PRODUCTS & 21 & 3 \\
APPAREL, EXCEPT FUR AND LEATHER APPAREL & 27 & 3 \\
BASIC ALUMINIUM PRODUCTS AND SPECIAL ALLOYS & 18 & 4 \\
SERVICE OF PROVIDING WASHED AND STERILIZED REUSABLE SURGICAL DRAPES... & 29 & 4 \\
BOTTLED WATERS, NOT SWEETENED OR FLAVOURED & 12 & 4 \\
RARE-EARTH CONCENTRATES, OXIDES, METALS, AND MAGNETS... & 15 & 4 \\
TRANSPORT SERVICES & 26 & 4 \\
SWEATERS, JERSEYS, PULLOVERS, CARDIGANS, FLEECES... & 31 & 3 \\
RAW SUGAR, REFINED SUGAR, AND MOLASSES & 22 & 4 \\
ESCALATORS AND MOVING WALKS & 34 & 4 \\
FURNITURE, EXCEPT SEATS AND MATTRESSES & 17 & 4 \\
PROCESSED LIQUID MILK AND CREAM & 14 & 3 \\
BIRDS’ EGGS IN SHELL, FRESH & 22 & 2
\end{tabular}
\end{flushleft}
\caption{The full list of all product categories that were evaluated in this work (the product categories with in-depth evaluations shown are highlighted). The product categories have a wide range of the number of processes (ranging from 12 processes to 46), and with a wide spread of product specificity (as indicated by the length of the shortest UN-CPC code)} }\label{tab:product_overview}
\end{table*}

\subsection{Prompts Used for Baselines}
In this section, we provide the prompts used to produce our two baselines (LLMDirect and LLMExample). The prompt for LLMDirect takes the input of a product category name (``product\_name") and its corresponding description (``product\_description"). This prompt is shown in Figure \ref{fig:llmdirect}.

\begin{figure*}
\small{
\begin{verbatim}
    """You are an expert in Life Cycle Assessment (LCA) and Product Category Rules (PCR).
    
    Your goal is to give the Upstream, Core, and Downstream processes for a specific product, 
    define the edges that connect them in a directed process flow graph.
    
    The given product name is {product_name}
    The given product description is {product_description}

    Based on the product name and description, follow the instructions below to 
    create the process flow graph.
    1. List any upstream processes that are involved with creating the product. 
    Describe each process
    2. List any core processes that are involved with creating the product. 
    Describe each process
    3. List any downstream processes that are involved with creating 
    the product. Describe each process

    Based on the process descriptions above, 
    carefully follow the instructions below to create the directed process flow graph:
    1. Order each of the processes
    by the sequence in which they are done and create edges between each of these processes. 
    2. If one of the subprocesses described above is a subprocess of another process,
    indicate this by creating an edge from 
    the subprocess to the process (ex. subprocess --> process). 
    Indicate that it is a subprocess. 

    Be sure to include steps between the upstream, 
    
    core and downstream processeses (ex. a transformation 
    activity between an upstream and core process) 
    so that the graph is connected throughout. 

    4. Give a single JSON based 
    on the information given above in the exact format given below:

    {{
    "processes": {{
                    <process_name> : {{
                        "description": <process_description>,
                        "process_category": <list either upstream, core or downstream>,
                        "is_subprocess": <list either subprocess or process>,
                        "input_nodes": [ input_node_1, input_node_2 ...
                        ], 
                        "output_nodes": [output_node_1, output_node_2 ...], 
                        “reasoning:  < provide a detailed description of the rationale>
            
                    }},
                     <process_name_2> : {{
                        "description": <component_description>,
                        "process_category": <list either upstream, core or downstream>,
                        "is_subprocess": <list either subprocess or process>,
                        "input_nodes": [ input_node_1, input_node_2 ...
                        ], 
                        "output_nodes": [output_node_1, output_node_2 ...], 
                        “reasoning:  < provide a detailed description of the rationale>
            
                    }},
                    ...
                }}
    }}
    Important Instructions:
    1. Ensure that the JSON format given is followed exactly.
    Do not follow any other JSON format
    2. Ensure your response includes a clear and concise breakdown of each process,
    using the information provided in the input JSON. 
    3. Be sure that each  process is as detailed as possible
    5. Provide details of all assumptions made and rationale behind each determination."""

\end{verbatim}
}

\caption{Prompt used for LLMDirect}\label{fig:llmdirect}
\end{figure*}

The prompt for LLMExample takes the input of a product category name (``product\_name") and its corresponding description (``product\_description"). It takes an additional input of an example \graphname (in the case of our implementation, the additional example was the \graphname for ``Baked Goods"). This prompt is shown in Figure \ref{fig:llmexample}. 
\begin{figure*}
\small{
\begin{verbatim}
    """You are an expert in Life Cycle Assessment (LCA) and Product Category Rules (PCR).
    Your goal is to give the Upstream, Core, and Downstream processes for a specific product,
    define the edges that connect them in a directed process flow graph.
    
    The given product name is {product_name}
    The given product description is {product_description}


    Below is an example PCR. Generate your response such that it 
    emulates the life cycle phases, and scopes of this document
    The document is the following, which describes the PCR for baked goods:
    {example}

    Based on the product name and description, follow the 
    instructions below to create the process flow graph.
    1. List any upstream processes that are involved with
    creating the product.
    Describe each process
    2. List any core processes that are involved with creating the product.
    Describe each process
    3. List any downstream processes that are involved with creating the product. 
    Describe each process

    Based on the process descriptions above, carefully follow the 
    instructions below to create the directed process flow graph:
    1. Order each of the processes by the sequence in which they are done and 
    create edges between each of 
    these processes. 
    2. If one of the subprocesses described above is a subprocess of another process,
    indicate this by creating an edge from the
    subprocess to the process (ex. subprocess --> process). Indicate that it is a subprocess. 

    Be sure to include steps between the upstream, core and downstream processeses 
    (ex. a transformation activity between an upstream and core process) 
    so that the graph is connected throughout. 

    4. Give a single JSON based on the information given above in the exact format given below:

    {{
    "processes": {{
                    <process_name> : {{
                        "description": <process_description>,
                        "process_category": <list either upstream, core or downstream>,
                        "is_subprocess": <list either subprocess or process>,
                        "input_nodes": [ input_node_1, input_node_2 ...
                        ], 
                        "output_nodes": [output_node_1, output_node_2 ...], 
                        ``reasoning":  < provide a detailed description of the rationale>
            
                    }},
                     <process_name_2> : {{
                        "description": <component_description>,
                        "process_category": <list either upstream, core or downstream>,
                        "is_subprocess": <list either subprocess or process>,
                        "input_nodes": [ input_node_1, input_node_2 ...
                        ], 
                        "output_nodes": [output_node_1, output_node_2 ...], 
                        ``reasoning":  < provide a detailed description of the rationale>
            
                    }},
                    ...
                }}
    }}
    
    Important Instructions:
    1. Ensure that the JSON format given is followed exactly. Do not follow any other JSON format
    2. Ensure your response includes a clear and concise breakdown of each process,
    using the information provided in the input JSON. 
    3. Be sure that each process is as detailed as possible
    5. Provide details of all assumptions made and rationale behind each determination."""
\end{verbatim}
}

\caption{Prompt used for LLMExample}\label{fig:llmexample}
\end{figure*}

\end{document}